\pdfoutput=1
\documentclass[11pt]{article}

\usepackage[table]{xcolor}
\usepackage[]{coling}
\usepackage{framed}
\usepackage{cancel}
\usepackage{graphicx}
\usepackage{subcaption}
\usepackage{booktabs}
\usepackage{tabularx}
\usepackage{multirow} % Required for multirow command
\usepackage{longtable}
\usepackage[table]{xcolor}
\usepackage{adjustbox}
\usepackage{colortbl}

\usepackage{mdframed}

\usepackage{times}
\usepackage{latexsym}
\usepackage{booktabs}
\usepackage{amsmath}
\usepackage{svg}

\usepackage{graphicx}
\usepackage{booktabs} % For professional-looking tables
\usepackage{caption} % For captions
\usepackage{array} % For column spacing
\usepackage[T1]{fontenc}
\usepackage[utf8]{inputenc}

\usepackage{microtype}

 \usepackage{todonotes}

\usepackage{comment}

\usepackage{xspace}
\title{%On 
Biases in Large Language Model-Elicited Text:\\A Case Study in Natural Language Inference}

\author{Grace Proebsting \\ Haverford College \\
  \texttt{gproebstin@haverford.edu} \\ \And
  Adam Poliak \\ Bryn Mawr College \\
  \texttt{apoliak@brynmawr.edu} \\ }

\begin{document}
\maketitle
\begin{abstract}
We test whether NLP datasets created with Large Language Models (LLMs) contain annotation artifacts and social biases like NLP datasets elicited from crowd-source workers.
We recreate a portion of the Stanford Natural Language Inference corpus using GPT-4, Llama-2 70b for Chat, and Mistral 7b Instruct. We train hypothesis-only classifiers to determine whether LLM-elicited NLI datasets contain annotation artifacts. Next, we use point-wise mutual information to identify the words in each dataset that are associated with gender, race, and age-related terms.
On our LLM-generated NLI datasets, fine-tuned BERT hypothesis-only classifiers achieve between 86-96\% accuracy.
Our analyses further characterize the annotation artifacts and stereotypical biases in LLM-generated datasets.
\end{abstract}

\section{Introduction}
Creating NLP datasets with Large Language Models (LLMs) is an attractive alternative to relying on crowd-source  workers~\cite{ziems2023large}.
Compared to crowd-source workers, LLMs are inexpensive, fast, and always available. Although LLMs require validation~\cite{pangakis2023automated}, they are an efficient tool to annotate data~\cite{zhao-etal-2022-lmturk,bansal2023large,gilardi2023chatgpt,he2023annollm}. %,ziems2023large}.
In addition to relying on LLMs for data annotation, researchers can elicit text from LLMs to create NLP datasets. For instance, LLMs have been used to generate training sets for NLP classification tasks like sentiment and intent classification~\cite{ye-etal-2022-zerogen,sahu-etal-2022-data,chung-etal-2023-increasing}.

Eliciting text from humans can yield NLP datasets with stereotypical biases~\cite{rudinger-etal-2017-social} and annotation artifacts~\cite{cai-etal-2017-pay,kaushik-lipton-2018-much}.
Since researchers use LLMs to create textual datasets, we study whether LLM-elicited datasets similarly suffer from stereotypical biases and annotation artifacts. To compare human- and machine-elicited textual data, we create LLM-generated versions of the Stanford Natural Language Inference (SNLI) corpus~\cite{bowman-etal-2015-large} by providing LLMs with the same instructions given to SNLI crowd-source workers. 

We focus on Natural Language Inference (NLI), the task of determining whether a hypothesis sentence could be likely inferred from a premise~\cite{10.1007/11736790_9}, since popular NLI datasets with crowd-sourced hypotheses contain biases. We apply standard approaches to detect annotation artifacts in NLI by training hypothesis-only classifiers and identifying %give-away words, i.e. 
words highly associated with specific NLI labels.
Further, we search for race, age, and gender-based stereotypical biases by finding words most associated with these social groups, and compare them with biases in SNLI.

We find that LLM-elicited NLI contains both hypothesis-only and social biases. On our LLM-generated NLI datasets, fine-tuned BERT classifiers achieve 86-96\% accuracy when given only the hypotheses, compared to 72\% performance on SNLI. We also find the LLM-generated datasets contain similar gender stereotypes as SNLI. Our research suggests that while eliciting text from LLMs to generate NLP datasets is enticing and promising, thorough quality control is necessary.

\section{Background \& Motivation}

\begin{table*}[t!]
\centering
%\small
{
\begin{tabular}{llll} \toprule
\textbf{Premise} & Two women are hiking in the wilderness. & \\\midrule
& \textbf{Entailment} & \textbf{Contradiction} \\
\textbf{SNLI} & There are two women outdoors. &  There are two women in the living room. \\ 
\textbf{Llama} & There are people outdoors. & A couple is having a picnic in a park.\\
\textbf{Mistral} & There are people in nature. & The women are shopping for clothes. \\
\textbf{GPT-4} & People are outdoors. & Two women are swimming in a pool. \\\bottomrule
\end{tabular}
}
\caption{Entailed and contradicted hypotheses produced by humans (SNLI) and three LLMs (Llama-2 70b for Chat, Mistral 7b Instruct, and GPT-4) in response to the same premise.}
\label{tab:example}
\end{table*}

There is a robust literature focusing on whether LLMs contain biases~\cite{nozza-etal-2021-honest,sheng-etal-2021-societal, 10.1145/3593013.3594109,kolisko2023exploring,10.1162/coli_a_00524,liu-etal-2024-evaluating-large,shin-etal-2024-ask,raj2024breaking,hu2024generative}. We similarly evaluate biases in LLMs, but our focus is different: specifically, we ask whether LLMs are a suitable replacement for crowdsource workers when creating NLP datasets. Concretely, we investigate whether NLP datasets with LLM-elicited text contain similar annotation artifacts and social biases as NLP datasets with human-elicited text.

Prompting humans to generate text for large-scale NLP datasets can lead to biased datasets. Famously, datasets for the Story Cloze Test and NLI contain biases introduced by their human elicitation protocols. 
To create a dataset for the Story Cloze Test, i.e. the task of determining the correct ending of a story, \newcite{mostafazadeh-etal-2016-corpus} asked crowd-source workers ``to write novel five-sentence stories.'' \newcite{bowman-etal-2015-large} created SNLI by providing crowd-source workers image captions from the Flickr30k corpus~\cite{young-etal-2014-image} and instructing workers to write three alternative captions: one that is \textit{definitely true}, one that \textit{might be true}, and one that is \textit{definitely false}. These human-elicitation protocols are responsible for creating 1) annotation artifacts that enable naive models ignoring substantial context to perform surprisingly well~\cite{schwartz-etal-2017-effect,tsuchiya-2018-performance,gururangan-etal-2018-annotation,poliak-etal-2018-hypothesis,feng-etal-2019-misleading}, and 2) social biases that ``amplify $\ldots$ stereotypical associations''~\cite{rudinger-etal-2017-social}.

In addition to these concerns, creating datasets by eliciting text from humans can be expensive. LLMs can efficiently generate, label, and clean datasets for a wide variety of applications~\cite{ziems2023large}. LLMs have been used to generate instruction-tuning datasets \cite{honovich-etal-2023-unnatural,wang-etal-2023-self-instruct, peng2023instruction}, synthetic versions of benchmarks like SuperGLUE~\cite{wang2020superglue, gupta2023targen}, counterfactuals for dataset augmentation~\cite{wu-etal-2021-polyjuice,chen-etal-2023-disco}, attributable information seeking~\cite{kamalloo2023hagrid}, and free-text classification explanations ~\cite{wiegreffe-etal-2022-reframing}. LLM-elicitation is especially attractive for sensitive domains, e.g. clinical NLP,  where datasets must not leak personal
identifying information~\cite{frei2023annotated,xu2023knowledgeinfused}.
LLMs-elicited text is pervasive even among crowd-source workers: ~\newcite{veselovsky2023artificial} claim that ``33–46\%" of the crowd-source workers hired for a summarization task likely used LLMs to produce summaries.

Some LLM-generated datasets involve no post-filtering step~\cite{peng2023instruction,xu2023contrastive,xu2023knowledgeinfused}. However, most resources built with LLM-elicitation include thorough quality assurance, either through ``human-in-the-loop'' curation ~\cite{wiegreffe-etal-2022-reframing,liu-etal-2022-wanli,kamalloo2023hagrid}, statistical filtering~\cite{wu-etal-2021-polyjuice,ye-etal-2022-zerogen,wang-etal-2023-self-instruct} or relying on neural models to filter LLM-generated data~\cite{wiegreffe-etal-2022-reframing,chen-etal-2023-disco,yehudai2024genie,gupta2023targen}. While we advocate for filtering steps to ensure quality and remove biases in LLM-elicited text, we focus on analyzing the unfiltered output of ``out-of-the-box'' LLMs for NLP datasets. We ask, specifically in the context of NLI, whether LLM-elicited text contains biases, and if so, what are these biases?

\section{Creating LLM-Elicited NLI}
We use NLI as a case study to explore whether LLM-generated text contain similar biases as human-written text
 since human-elicited NLI 
datasets contain annotation artifacts and stereotypical social biases.
We create modified versions of SNLI by prompting LLMs with the same instructions that \newcite{bowman-etal-2015-large} gave to crowd-source workers.
\autoref{tab:example} provides examples from each dataset.
We further verify the quality of the generated hypotheses and determine how different they are from those in SNLI.

\paragraph{LLMs under consideration}
\label{LLMs}

We select a diverse set of LLMs for dataset generation: \textbf{GPT-4} \cite{openai2023gpt4}, \textbf{Llama-2 70b for Chat} \cite{touvron2023llama}, \textbf{Mistral 7b Instruct} \cite{jiang2023mistral}, and \textbf{PaLM 2 for Chat} \cite{anil2023palm}. \footnote{For GPT-4 we use \texttt{gpt-4-0613}, for Llama Chat 70b we use \texttt{llama-2-70b-chat}, for Mistral 7b Instruct we use \texttt{mistral-7b-instruct-v0.1}, and for PaLM 2 for Chat we use \texttt{chat-bison}.} 
 These models vary in parameter count, parent company, and training technique.
 
We initially included models with open training sets to test for data contamination, e.g. AI2's OLMo-7B-Instruct~\cite{groeneveld2024olmo}, DataBrick's dolly-v2-12b~\cite{DatabricksBlog2023DollyV2} or EleutherAI’s gpt-j-6b~\cite{wang2021gpt}, but these open-data models did not create accurate entailed hypotheses in initial experiments. Given computational constraints, we were unable to use LLMs, e.g. BLOOM~\cite{workshop2022bloom} or Falcon~\cite{almazrouei2023falcon}.

\begin{table}[t!]
\center
  \begin{tabular}{l r} 
    \toprule
\multicolumn{2}{l}{\textbf{Data set sizes:}}\\
Training pairs &  133,629\\
Evaluation pairs &  6,525\\
\midrule
\multicolumn{2}{l}{\textbf{Hypothesis mean token count:}}\\
SNLI train & 8.1 \\
Llama train & 9.4 \\
Mistral train & 9.1 \\
GPT-4 train & 9.2 \\
PaLM 2 train & 7.7 \\
\midrule
\multicolumn{2}{l}{\textbf{Mean Jaccard similarity with SNLI:}}\\
Llama train & 0.19 \\
Mistral train & 0.22 \\
GPT-4 train & 0.20 \\
PaLM 2 train & 0.25 \\
    \bottomrule
  \end{tabular}
\caption{Summary statistics for each dataset.}
\label{tab:collection-stats}
\end{table}

\paragraph{Dataset generation}
To mirror \newcite{bowman-etal-2015-large}'s dataset elicitation pipeline, we prompted LLMs with the same instructions provided to crowd-source workers for SNLI.\footnote{We slightly changed the prompt to ensure the LLM's output was valid JSON. We provide the full prompt in the Appendix (\autoref{instructions-1}).} 

To balance lexical diversity with reproducibility, we set the temperature and top-p respectively to $0.75$ and $0.9$ for all LLMs. Additionally, we use the default top-k parameter for each LLM. Due to budget constraints, for each LLM, we create hypotheses for a third of the premises in the SNLI train set and all premises in the SNLI evaluation set. \autoref{tab:collection-stats} contains statistics regarding each dataset.

\begin{table}[h!]
    \centering
    \begin{tabular}{c|c|c|c|c}
    \toprule
         &  Overall&  Entail&  Neutral& Contra\\ \midrule
         SNLI&  92.7&  87.0&  95.0& 96.0\\
         Llama &  89.7&  73.0&  98.0& 98.0\\
         Mistral &  83.7&  70.0&  91.0& 90.0\\
         GPT-4 &  94.3&  84.0&  99.0& 100.0\\
         PaLM 2&  77.0&  62.0&  90.0& 79.0\\ \bottomrule
    \end{tabular}
    \caption{Percentage of examples where we agreed with the label of 300 NLI example pairs from each dataset.} 
    \label{tab:validation}
\end{table}

\begin{figure}
    \centering
    \includegraphics[width=1\linewidth]{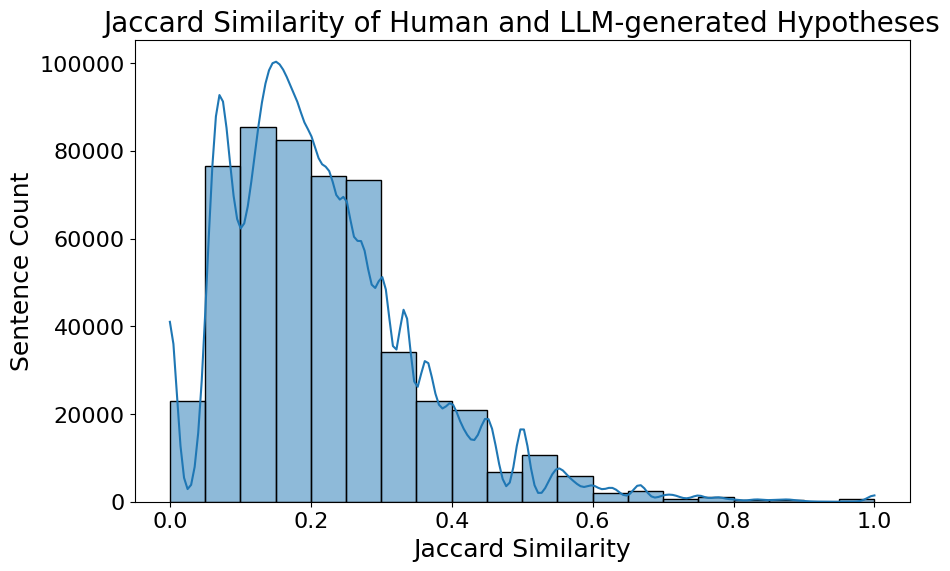}
    \caption{Frequency (y-axis) of lexical overlap (x-axis) between LLM and corresponding SNLI hypotheses. } 
    \label{fig:jaccard}
\end{figure}

\begin{figure*}[t!]
    \centering
    \includegraphics[width=0.9\linewidth]{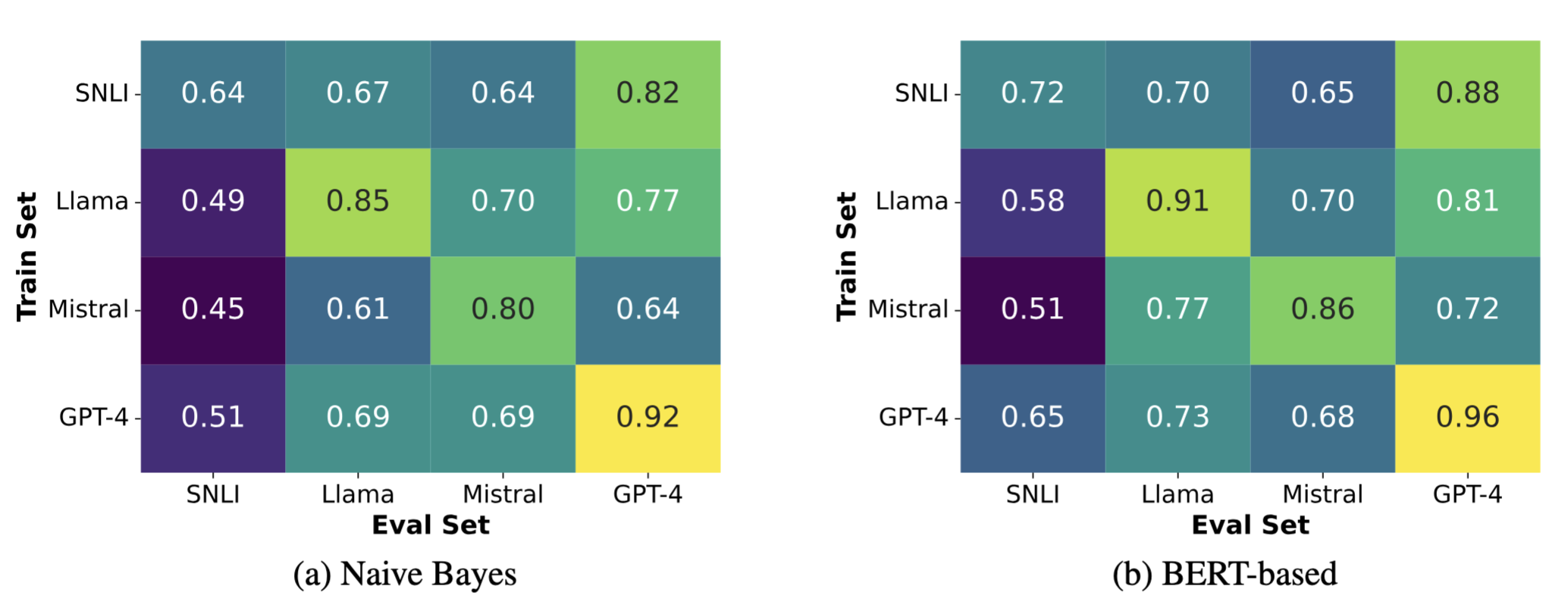}
    % \begin{subfigure}[t]{0.45\textwidth}
    % \centering
    % \includesvg[width=200pt]{naive_bayes_heatmap_v999.svg}
    %     \caption{Naive Bayes}\label{fig:naive_bayes}
    % \end{subfigure}%
    % ~ 
    % \begin{subfigure}[t]{0.45\textwidth}
    % \centering
    % \includesvg[width=200pt]{bert_heatmap_v999.svg}
    %     \caption{BERT-based} \label{fig:bert}
    % \end{subfigure}
    \caption{Accuracy of each hypothesis-only classifier on each LLM and human-generated evaluation set. Each row represents the hypothesis-only NLI dataset used for training, and each column represents the evaluation dataset.}
    \label{fig:hyp-only-llms}
\end{figure*}

\paragraph{Dataset validation}

To verify the LLMs correctly generated hypotheses for each label, we sampled 100 premises and manually verified the labels for the corresponding 300 NLI sentence pairs for each model. \autoref{tab:validation} reports our agreement with the NLI labels for each LLM. Since we agreed with less than 80\% of the examples sampled from the PaLM2-elicited dataset, we do not consider the dataset generated by PaLM2 in our later studies.

To ensure the LLM-generated hypotheses are not simply memorized and copied verbatim from SNLI, we compute the Jaccard similarity of the words within pairs of LLM-generated and SNLI hypotheses corresponding to the same premises and labels.\footnote{Jaccard similarity is a measure of set overlap that ranges between 0.0 (a disjoint set) and 1.0 (an identical set).}

\autoref{fig:jaccard} plots the distribution of the Jaccard similarities between  SNLI and corresponding LLM-generated hypotheses. \autoref{tab:collection-stats} reports the average Jaccard similarity for each \textit{individual} LLM dataset. \textbf{LLM and human-generated hypotheses have low lexical overlap}, demonstrating that these LLMs do not copy SNLI verbatim.\footnote{Reviewers noted the limits of Jaccard similarity since LLMs might paraphrase hypotheses from SNLI if the LLMs were pre-trained on SNLI.
 A manual review of thousands of examples suggested that these LLM-generated hypotheses contained semantically different content from that of the hypotheses in SNLI, i.e., the LLM-generated hypotheses were not merely paraphrased from SNLI.}

\section{Study 1: Hypothesis-Only Artifacts}
In our first study, we determine whether LLM-elicited NLI datasets contain annotation artifacts that allow hypothesis-only models to outperform a majority-class baseline. We train two types of hypothesis-only models: Naive Bayes (NB) using the case-sensitive implementation from scikit-learn with unigram features~\cite{scikit-learn}, and a fine-tuned BERT classifier~\cite{DBLP:journals/corr/abs-1810-04805}, specifically bert-base-uncased models with 3-class sequence classification heads and default HuggingFace hyper-parameters~\cite{wolf-etal-2020-transformers},\footnote{We did not perform hyper-parameter tuning since our goal is simply to establish whether a hypothesis-only model can perform well on an LLM-elicited NLI dataset.} which we train for 1 epoch using AdamW \cite{loshchilov2019decoupled}, a learning rate of 2e-5, a weight decay of 0.01, and a batch size of 16.

We train hypothesis-only models on each of our train sets (3 LLM-generated and the filtered SNLI) and evaluate them on all evaluation sets.
\autoref{fig:hyp-only-llms}
reports the accuracy of the hypothesis-only models.

The highest-performing model on each evaluation set was trained on the corresponding train set - in each column in \autoref{fig:hyp-only-llms}, the highest accuracy is along the diagonal.

Surprisingly, the SNLI-trained models perform much better on the GPT-4 generated evaluation set (0.82 for NB and 0.88 for BERT) than on the SNLI evaluation set (0.64 for NB and 0.72 for BERT), indicating that GPT-4 might contain similar annotation artifacts as SNLI.

\begin{table*}[bth!]
%\small
\begin{adjustbox}{width=\textwidth}

\centering
\subfloat[][entailment]{
\begin{tabular}[h]{l|c c c}

\toprule
& \textbf{Word} & \textbf{$p(l|w)$} & \textbf{Freq} \\
\midrule

& Humans & 0.95 & 128 \\
& least & 0.92 & 78 \\
& activity & 0.83 & 47 \\
& multiple & 0.81 & 37 \\
& interacting & 0.85 & 34 \\
SNLI & motion & 0.97 & 32 \\
& physical & 0.83 & 30 \\
& occupied & 0.8 & 15 \\
& balances & 0.82 & 11 \\
& consuming & 0.8 & 10 \\
\midrule 
& person & 0.81 & 22264 \\
& People & 0.86 & 7059 \\
& standing & 0.84 & 4359 \\
& outdoors & 0.93 & 2390 \\
& engaging & 0.94 & 1689 \\
Llama & Three & 0.92 & 1593 \\
& gathered & 0.93 & 1513 \\
& activity & 0.83 & 1412 \\
& public & 0.82 & 1230 \\
& vehicle & 0.87 & 1185 \\
\midrule

& There & 0.99 & 16707 \\
& outdoors & 0.87 & 1055 \\
& three & 0.83 & 720 \\
& four & 0.88 & 335 \\
& urban & 0.83 & 318 \\
Mistral & consuming & 0.94 & 217 \\
& multiple & 0.83 & 211 \\
& vertical & 0.84 & 182 \\
& acrobatic & 0.88 & 176 \\
& many & 0.87 & 153 \\
\midrule
& person & 0.85 & 11764 \\
& outdoors & 0.97 & 8182 \\
& individual & 0.96 & 4569 \\
& individuals & 0.89 & 3878 \\
& There & 0.86 & 3794 \\
GPT-4 & Individuals & 0.97 & 2159 \\
& interacting & 0.98 & 1377 \\
& activity & 0.97 & 1250 \\
& gathered & 0.88 & 1248 \\
& public & 0.85 & 976 \\
\bottomrule
\end{tabular}
}
\subfloat[][neutral]{
\begin{tabular}[h]{c c c}

\toprule
\textbf{Word} & \textbf{$p(l|w)$} & \textbf{Freq} \\
\midrule

tall & 0.85 & 418 \\
sad & 0.81 & 322 \\
first & 0.87 & 298 \\
owner & 0.83 & 284 \\
birthday & 0.83 & 227 \\
winning & 0.88 & 186 \\
favorite & 0.88 & 180 \\
professional & 0.83 & 149 \\
vacation & 0.94 & 141 \\
win & 0.86 & 140 \\
\midrule
Someone & 1 & 4092 \\
trying & 0.9 & 3023 \\
going & 0.95 & 1604 \\
break & 0.87 & 1339 \\
fun & 0.88 & 1165 \\
practicing & 0.86 & 1142 \\
ride & 0.82 & 811 \\
or & 0.83 & 795 \\
discussing & 0.88 & 720 \\
catch & 0.95 & 622 \\
\midrule

be & 0.97 & 5154 \\
trying & 0.8 & 4875 \\
may & 0.98 & 3815 \\
having & 0.85 & 2039 \\
going & 0.83 & 1877 \\
or & 0.86 & 1858 \\
friends & 0.95 & 1499 \\
It & 0.9 & 1486 \\
could & 0.98 & 1311 \\
fun & 0.92 & 1201 \\
\midrule

to & 0.85 & 7087 \\
for & 0.89 & 5791 \\
his & 0.82 & 5042 \\
friends & 0.94 & 3439 \\
enjoying & 0.85 & 2073 \\
couple & 0.81 & 1878 \\
from & 0.82 & 1823 \\
taking & 0.82 & 1093 \\
practicing & 0.87 & 1092 \\
team & 0.88 & 972 \\
\bottomrule
\end{tabular}
}
\subfloat[][contradiction]{
\begin{tabular}[h]{c c c}

\toprule
\textbf{Word} & \textbf{$p(l|w)$} & \textbf{Freq} \\
\midrule

sleeping & 0.84 & 1747 \\
Nobody & 0.93 & 592 \\
asleep & 0.83 & 523 \\
couch & 0.81 & 477 \\
naked & 0.88 & 248 \\
tv & 0.81 & 207 \\
cats & 0.89 & 199 \\
TV & 0.81 & 177 \\
No & 0.93 & 134 \\
television & 0.83 & 124 \\
\midrule
celebrity & 0.92 & 2359 \\
actually & 0.94 & 2075 \\
cat & 0.9 & 1973 \\
Everyone & 0.93 & 1913 \\
adult & 0.89 & 1782 \\
fashion & 0.85 & 1766 \\
red & 0.84 & 1537 \\
signing & 0.92 & 1437 \\
autographs & 0.93 & 1398 \\
sleeping & 0.82 & 1371 \\
\midrule

The & 0.81 & 38491 \\
sitting & 0.83 & 14564 \\
bench & 0.87 & 8545 \\
not & 0.94 & 8068 \\
subject & 0.87 & 3672 \\
couch & 0.91 & 2330 \\
empty & 0.89 & 1433 \\
cards & 0.92 & 1171 \\
no & 0.92 & 955 \\
movie & 0.9 & 938 \\
\midrule
swimming & 0.92 & 16281 \\
pool & 0.91 & 14638 \\
reading & 0.8 & 3492 \\
book & 0.81 & 3048 \\
sleeping & 0.91 & 2326 \\
cooking & 0.84 & 2126 \\
cat & 0.9 & 1875 \\
dress & 0.8 & 1537 \\
alone & 0.94 & 1293 \\
library & 0.91 & 1274 \\
\bottomrule
\end{tabular}
}
\end{adjustbox}
\caption{The most highly correlated words for each train set for given labels (the columns (c), (d), and (e)), thresholded to those with $p(l|w) >= 0.8$ and ranked according to frequency.}\label{snli}
\end{table*}

We also notice that hypothesis-only models trained on LLM-generated data perform much better on other LLM-elicited datasets than on SNLI, as the accuracies in the first column are much lower than the other columns in both figures. This might indicate that the LLMs produce similar biases.

\paragraph{Qualitative analysis of give-away words}
The NB models with unigram features significantly outperform a majority baseline (\autoref{fig:hyp-only-llms}), indicating that the hypotheses contain \textit{give-away words}---single words that are highly indicative of a label.

We identify give-away words for each train set by calculating the conditional probability of each label $l$ given the presence of a word $w$ in a hypothesis:
$p(l|w) = \dfrac{count(w,l)}{count(w)}$.
We consider all give-away words with a conditional probability of at least 0.8. We follow \newcite{poliak-etal-2018-hypothesis} and sort give-away words by their frequency ``since
this statistic is perhaps more indicative of a word
$w$’s effect on overall performance compared to
$p(l|w)$ alone.'' ~\autoref{snli} reports the top $10$ give-away words for each label in all train sets.

%\paragraph{Entailment.} 
%Human-generated entailments
Entailed examples in SNLI often contain generic words like \textit{humans}, \textit{activity}, and \textit{interacting}. We find a similar pattern in LLM-generated entailed hypotheses, e.g. \textit{person} and \textit{activity} in GPT-4 and Llama. Unlike in SNLI, the capitalized word \textit{There} is a give-away for LLM-elicited entailed examples. 

LLMs often copy features from examples in prompts~\cite{elhage2021mathematical, olsson2022context, bansal-etal-2023-rethinking,zhang2023benchmarking}, which might explain why \textit{There} is a give-away word in these LLM-elicited datasets. 

Human-generated neutral hypotheses often contain modifiers (\textit{tall, sad, professional}) and superlatives (\textit{first, favorite, winning}). LLMs similarly add embellishing details about emotions or intentions (\textit{enjoying, fun, practicing, trying}) or the relationships between agents (\textit{friends, couple, team}) that are not explicit in the premise. %Notably, t
Two of Llama's neutral give-away words, \textit{Someone} and \textit{catch}, appear in the prompt's example of a neutral hypothesis.

Lastly, both human- and LLM-elicited contradicting hypotheses contain % Human-generated contradictions often
negation words, e.g. \textit{nobody}, \textit{no}, \textit{not}.  
As noted by \newcite{poliak-etal-2018-hypothesis}, premises ``sourced from Flickr naturally deal with activities.'' Therefore, similar to how contradicted hypotheses in SNLI often mention \textit{sleeping}, it is not surprising that LLM-elicited contradictions mention 
actions that cannot occur simultaneously to the action in the premise, e.g. \textit{swimming} for GPT-4 and \textit{sitting} for Mistral. 
Further, these verbs often occur in frequently repeated phrases that negate an action described in the premise. For example, the phrases \textit{``swimming in a pool''} and \textit{``sitting on a bench''} respectively occur more than 10,000 times in the GPT-4 and Mistral-generated train sets. 

\begin{figure}[t!]
    \centering
     \includegraphics[width=\columnwidth]{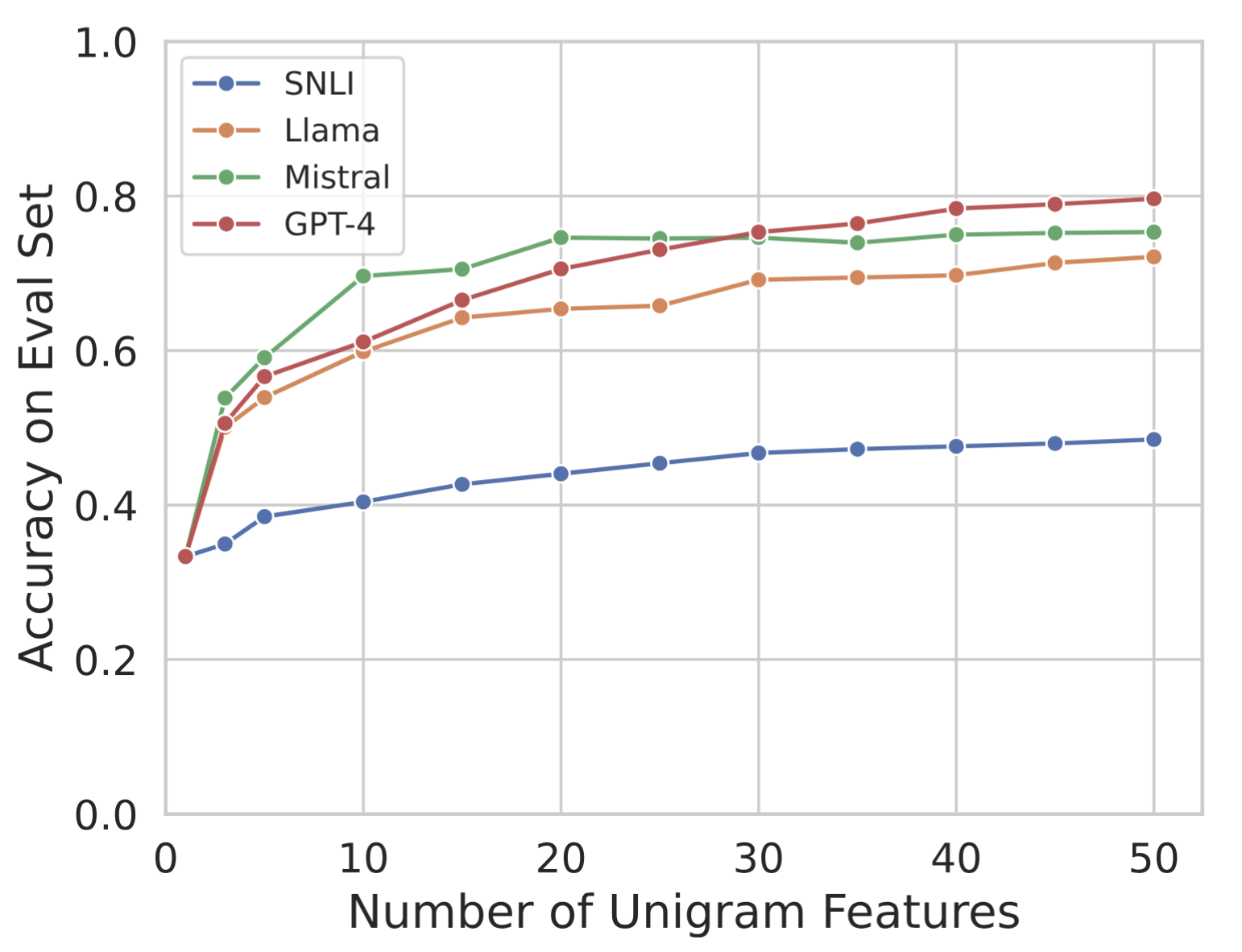}
    \caption{Accuracy of NB models using only the $n$ "most informative" unigram features for each train set evaluated on its corresponding evaluation set.}
    \label{fig:variable-features}
\end{figure}

\paragraph{Few unigrams needed for high NB accuracy.}

How many give-away words are necessary to accurately classify LLM-elicited NLI? To study this question, we train NB models that \textit{only receive the \textbf{n} most informative give-away words as features.} We find the most informative words for each train set by performing a chi-squared test on all words with respect to each label. We threshold to the top $n$ most informative unigrams and use only these words to train each $n$-feature NB model. \autoref{fig:variable-features} reports the accuracy of NB hypothesis-only models using just 1 to 50 features.

Compared to SNLI, the LLM-elicited datasets are far easier to classify using a sparse selection of unigram features. For example, with just 10 unigrams, all LLM-trained NB models achieve greater than 60\% accuracy, while the SNLI-trained 10-feature NB model only narrowly outperforms the majority-class baseline. This result indicates that LLM-generated hypotheses are trivial to classify not only due to the simplicity of the necessary features (unigrams) but also because only a negligibly small number of these simple features are required.

\begin{figure}[t!]
    \centering
  \includegraphics[width=\columnwidth]{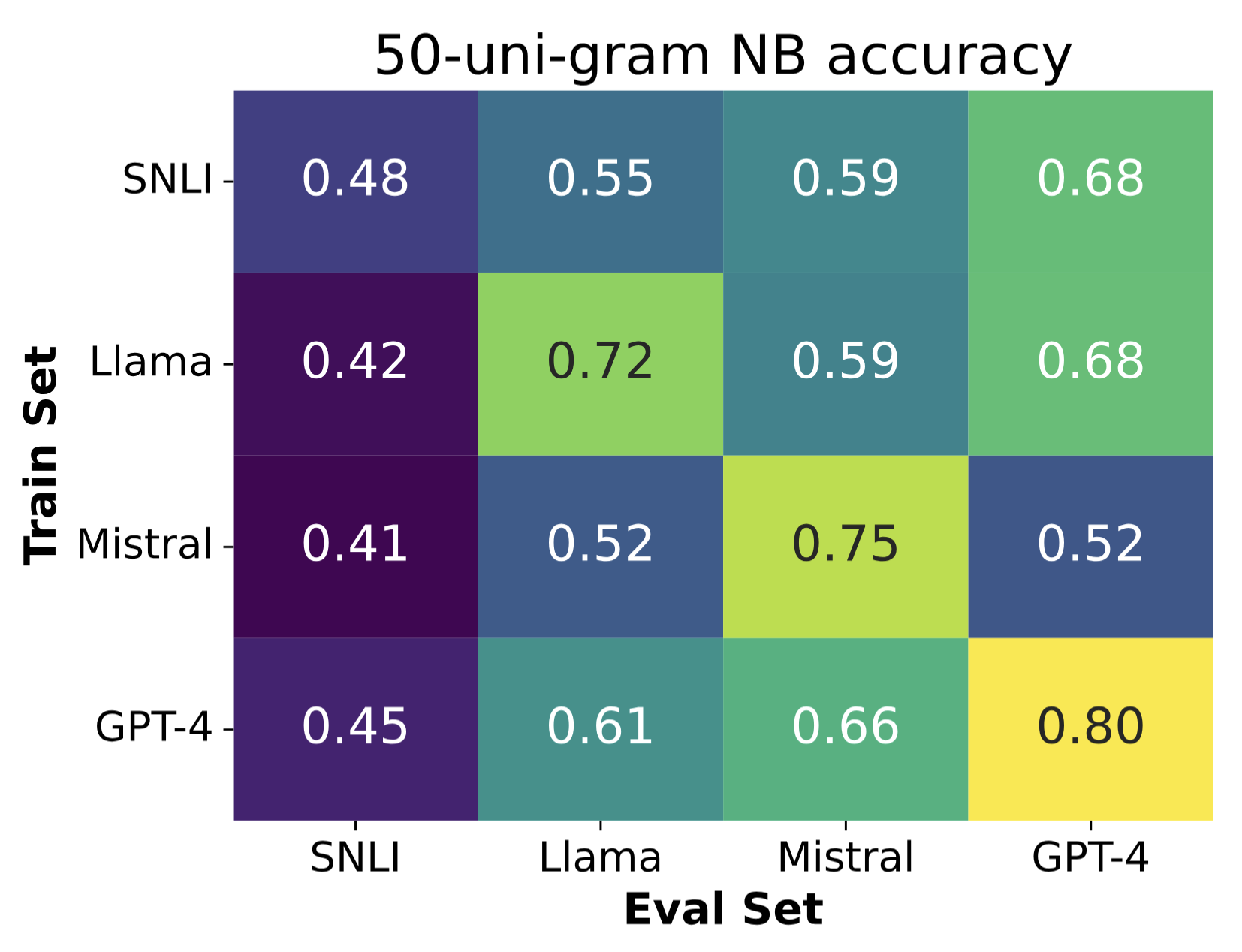}
    \caption{Accuracy of NB models with only the fifty most informative unigram features from their train set.}
    \label{fifty-feature-accuracy}
\end{figure}

\autoref{fifty-feature-accuracy} reports the accuracy of 50-unigram--feature NB models when evaluated on all four evaluation sets.
NB models trained with sparse unigram feature sets on the LLM-generated hypotheses outperform a random baseline on the evaluation sets of the other LLM-generated hypotheses. This suggests that highly informative unigram features from one LLM-elicited dataset can be informative on the other LLM-elicited datasets. Additionally, like the NB and BERT-based hypothesis-only models trained on the entire feature set, the 50-feature NB hypothesis-only model trained on SNLI performs better on the GPT-4 evaluation than the SNLI evaluation set. Overall, these results suggest that the high accuracy of full-feature NB models across the evaluation sets might be attributed to a sparse set of give-away words that are common across the LLM-elicited datasets. 

\section{Study 2: Stereotypical Biases}

Our second study analyzes whether LLM-elicited versions of SNLI, like the human-elicited SNLI, contain stereotypical social biases.
Following \newcite{rudinger-etal-2017-social}, we use pointwise mutual information (PMI) to identify words in each dataset that are most associated with gendered, racial, or age-based terms. Given word $w_1$ and $w_2$, the PMI between $w_1$ and $w_2$ is $\log(\dfrac{p(w_1,w_2)}{p(w_1)p(w_2)})$. For each dataset, we find the top co-occurring words in hypotheses by PMI with race, gender, and age-related query words that co-occur at least 3 times. 

\begin{table*}[t!]
\begin{mdframed}[
  linewidth=1pt, % Adjust the line width as desired
  innerleftmargin=10pt, % Adjust the left margin as desired
  innerrightmargin=10pt, % Adjust the right margin as desired
  frametitlerule=true, % Add a rule between the frame title and content
  %frametitle=\textbf{Within-Hypothesis PMI} % Customize the frame title
]
%\small
\underline{\textbf{SNLI}}\\
\textbf{woman} mascara$^\dagger$ knits$^\ddagger$ applies$^\ddagger$ sleds lipstick makeup$^\ddagger$ secret knitting$^\ddagger$ scarf countryside \\
        \textbf{man} burns surfs nose buys container internet orders tractor popcorn dives \\
        \textbf{women} cakes$^\ddagger$ yoga praying volleyball dresses thinking fruit tea talking$^\ddagger$ dance \\
        \textbf{men} burn compete$^\dagger$ wrestling suits$^\dagger$ poker celebrate passing uniforms chess cars \\

\underline{\textbf{GPT-4}}\\
\textbf{woman} ballgown gala bikini$^\ddagger$ oven$^\ddagger$ dress$^\ddagger$ ballroom$^\ddagger$ cookies$^\ddagger$ baking$^\ddagger$ heels$^\ddagger$ button \\
        \textbf{man} spiderman shaving$^\ddagger$ suit$^\ddagger$ mowing$^\ddagger$ hamburger tuxedo beard$^\ddagger$ tie$^\ddagger$ proposing frowning$^\ddagger$ \\
        \textbf{women} dresses$^\ddagger$ mall$^\ddagger$ yoga$^\dagger$ tea shopping$^\ddagger$ relaxing$^\dagger$ picnic$^\ddagger$ sunbathing$^\ddagger$ baking dancing$^\dagger$ \\
        \textbf{men} suits$^\ddagger$ hats laying installing football$^\ddagger$ hard basketball$^\ddagger$ gym$^\ddagger$ rodeo skyscraper$^\ddagger$ \\

\underline{\textbf{Llama}}\\
\textbf{woman} lap$^\ddagger$ makeup$^\ddagger$ applying$^\ddagger$ arms$^\ddagger$ nails$^\ddagger$ mirror$^\ddagger$ sink knitting$^\ddagger$ sunbathing$^\ddagger$ flower$^\dagger$ \\
        \textbf{man} shaving$^\ddagger$ basketball$^\ddagger$ beard$^\ddagger$ guitar$^\ddagger$ girlfriend three golf$^\ddagger$ stadium$^\ddagger$ walks$^\ddagger$ ironing \\
        \textbf{women} tea$^\dagger$ clothing$^\ddagger$ socializing smiling each other routine party standing dancing \\
        \textbf{men} football$^\ddagger$ dark field$^\ddagger$ basketball$^\ddagger$ instruments games$^\ddagger$ video$^\ddagger$ inside room playing$^\ddagger$ \\

\underline{\textbf{Mistral}}\\
\textbf{woman} cradling$^\ddagger$ arms sewing baby$^\ddagger$ flower$^\dagger$ newborn serving gymnastics herself her$^\ddagger$ \\
        \textbf{man} diving$^\dagger$ thrown net western tame horse wild his$^\ddagger$ cutting swinging$^\ddagger$ \\
        \textbf{women} japanese$^\dagger$ traditional$^\dagger$ clothes talking groceries posing conversation shopping smiling relaxing \\
        \textbf{men} boxing$^\ddagger$ suits$^\ddagger$ robes ring sparring$^\dagger$ fighting$^\ddagger$ court football basketball$^\ddagger$ match \\
        
\end{mdframed}
\caption{Top-ten words in hypothesis by PMI with gender-related query words in the same hypothesis, filtered to co-occurrences of at least three. (Hypothesis words that also appear in the premise are not included.) Significance of a likelihood ratio test for independence denoted by $\dagger$ ($\alpha$ = 0.01) and $\ddagger$ ($\alpha$ = 0.001).}
  \label{gender-bias-pmi}
\end{table*}

\begin{table*}[t!]
\centering
\small
\definecolor{lightgray}{gray}{0.9}
\begin{tabularx}{\textwidth}{|l|X|X|X|}
\hline
Query & \textbf{ENTAILMENT} & \textbf{NEUTRAL} & \textbf{CONTRADICTION} \\
\hline
\rowcolor{lightgray}
\textbf{man} & \textbf{SNLI}: often gun climbs a$^\ddagger$ seated & \textbf{SNLI}: stops bald cowboy cafe newspaper & \textbf{SNLI}: gas scooter wife sings wears \\
 & \textbf{GPT-4}: bathroom firearm casual embracing machine & \textbf{GPT-4}: latte cigar warehouse guy$^\ddagger$ adventurer & \textbf{GPT-4}: café bikini hat dolphins formal \\
\rowcolor{lightgray}
 & \textbf{Llama}: entertaining his$^\ddagger$ paper wood father$^\ddagger$ & \textbf{Llama}: article summit fan avoid seafood & \textbf{Llama}: waters packed negotiating kidnapping before \\
 & \textbf{Mistral}: presentation romantic moment a$^\ddagger$ scaling & \textbf{Mistral}: conference debris board summit a$^\ddagger$ & \textbf{Mistral}: shirt costume tie a$^\ddagger$ individual$^\ddagger$ \\
\hline
\rowcolor{lightgray}
\textbf{men} & \textbf{SNLI}: workers guys$^\dagger$ ball several they & \textbf{SNLI}: businessmen$^\dagger$ crew workers$^\dagger$ charity construction$^\dagger$ & \textbf{SNLI}: ladies break party enjoying lunch \\
 & \textbf{GPT-4}: workers construction$^\ddagger$ machinery project site & \textbf{GPT-4}: guys$^\ddagger$ foundation industrial soldiers$^\ddagger$ workers$^\ddagger$ & \textbf{GPT-4}: individuals$^\ddagger$ playground$^\ddagger$ women$^\ddagger$ people$^\ddagger$ everyone$^\ddagger$ \\
\rowcolor{lightgray}
 & \textbf{Llama}: parade$^\dagger$ marching$^\ddagger$ industrial formal construction$^\ddagger$ & \textbf{Llama}: cowboys$^\ddagger$ soldiers$^\ddagger$ complex$^\dagger$ fishermen workers$^\ddagger$ & \textbf{Llama}: awards$^\dagger$ ballet celebrities$^\ddagger$ players$^\ddagger$ parade \\
 & \textbf{Mistral}: fishermen$^\ddagger$ workers$^\ddagger$ job$^\dagger$ military$^\dagger$ personnel & \textbf{Mistral}: workers$^\ddagger$ soldiers$^\ddagger$ cowboys long-distance vendors & \textbf{Mistral}: casual admiring dressed$^\ddagger$ they$^\ddagger$ already \\
\hline
\rowcolor{lightgray}
\textbf{woman} & \textbf{SNLI}: her$^\ddagger$ touching lady a$^\ddagger$ women & \textbf{SNLI}: herself husband$^\dagger$ dress won clothes & \textbf{SNLI}: feeding a$^\ddagger$ phone she nothing \\
 & \textbf{GPT-4}: female$^\ddagger$ stand exiting lady$^\ddagger$ toys & \textbf{GPT-4}: quilt$^\ddagger$ businesswoman bag lady$^\ddagger$ casual & \textbf{GPT-4}: lady$^\ddagger$ suit$^\ddagger$ man$^\ddagger$ a$^\ddagger$ dinner \\
\rowcolor{lightgray}
 & \textbf{Llama}: scientist mother$^\ddagger$ her customer off & \textbf{Llama}: savoring meditating furry considering hiker & \textbf{Llama}: perched premiere bicycle singing world \\
 & \textbf{Mistral}: exiting scientist her$^\ddagger$ speaking a$^\ddagger$ & \textbf{Mistral}: lady else beauty her$^\ddagger$ hands & \textbf{Mistral}: makeup accessories getting her shopping \\
\hline
\rowcolor{lightgray}
\textbf{women} & \textbf{SNLI}: ladies$^\dagger$ woman$^\ddagger$ performing a$^\ddagger$ group & \textbf{SNLI}: woman$^\ddagger$ party a$^\ddagger$ group tall & \textbf{SNLI}: lunch men$^\dagger$ they a$^\ddagger$ play \\
 & \textbf{GPT-4}: ladies$^\ddagger$ females$^\ddagger$ lady$^\ddagger$ conversation walking & \textbf{GPT-4}: ladies$^\ddagger$ fruits vegetables female$^\ddagger$ restaurant & \textbf{GPT-4}: suits$^\ddagger$ ladies$^\dagger$ men$^\ddagger$ meeting business \\
\rowcolor{lightgray}
 & \textbf{Llama}: costumes gathering sporting dancing socializing & \textbf{Llama}: ladies shopping choreographed store local & \textbf{Llama}: men$^\ddagger$ football celebrities$^\ddagger$ during competing \\
 & \textbf{Mistral}: athletes people$^\ddagger$ clothing street outdoors & \textbf{Mistral}: females$^\ddagger$ female$^\ddagger$ woman$^\ddagger$ singing show & \textbf{Mistral}: people$^\ddagger$ clothing being any performers \\
\hline
\end{tabularx}
\caption{Top-five words in hypotheses of a particular label by PMI with gender-related query words in the premise, filtered to co-occurrences of at least three. (Hypothesis words that also appear in the premise are not included.) Significance of a likelihood ratio test for independence denoted by $\dagger$ ($\alpha$ = 0.01) and $\ddagger$ ($\alpha$ = 0.001).}
\label{gender-bias-pmi-between-prem-hypo}
\end{table*}

\paragraph{Gender-based stereotypes.}
\autoref{gender-bias-pmi} reports the top PMI terms for \textit{man, men, woman} and \textit{women.} PMI results for all query words can be found in the Appendix.
In both the human-elicited and LLM-elicited datasets, male query words are associated with violence, work, and physical activity. In SNLI these terms include \textit{burns, surfs, compete, wrestling, suits, poker, uniforms, chess, cars}. In the LLM-elicited datasets, terms highly associated with male terms include \textit{suit, mowing, basketball, golf, cutting, boxing, sparring,} and \textit{fighting}.

In SNLI, the female query words are associated with physical appearance (\textit{mascara, lipstick, makeup, dresses}) and leisure activities (\textit{knits, yoga, cakes, tea, talking, dance}). LLM-generated hypotheses display similar stereotypes: female query words are related to domesticity (\textit{oven, cookies, baking, knitting, cradling, baby, sewing, groceries}) and leisure activities (\textit{mall, yoga, tea, shopping, relaxing, picnic, sunbathing, dancing, socializing, party, talking}). In the LLM-elicited datasets, female query words are also associated with clothing and physical appearance (\textit{bikini, dress, heels, lap, makeup, arms, nails, clothing}). 

\paragraph{Label-specific gender biases.}
To study how stereotypical biases appear based on NLI labels, for each NLI label, we now compute the PMI of hypothesis words with query words that appear in the premise. This allows us to determine if the LLMs contain stereotypical biases that are specific to different NLI labels.
\autoref{gender-bias-pmi-between-prem-hypo} reports label-specific biases for gender-related queries. 

Broadly, LLM-generated entailed and neutral hypotheses display similar biases as the overall PMI results: male query words are associated with violence, physicality, and work, e.g. \textit{workers, military, soldiers}, while female query words are associated with leisurely or domestic activities and physical appearance, e.g. \textit{quilt, party, beauty}. A notable exception is that both Llama and Mistral associate ``woman'' with \textit{scientist} and GPT-4 associates ``woman'' with \textit{businesswoman}.\footnote{Respectively entailment and neutral columns in \autoref{gender-bias-pmi-between-prem-hypo}.} Additionally, Llama and Mistral associate ``women'' with \textit{sporting} and \textit{athletes}, respectively.\footnote{Entailment column in \autoref{gender-bias-pmi-between-prem-hypo}.}

Both human and LLM-generated \textit{contradictions} sometimes flip the gender of the subject between the premise and hypothesis.  In SNLI contradictions, male premise words are associated with \textit{ladies} and \textit{wife}, and LLM-generated contradictions feature \textit{bikini} and \textit{women}. Similarly, female premise words are often associated with \textit{suit, man, men, football, meeting, competing, business}, which might demonstrate a gender bias.

\paragraph{Race \& age biases}
Unlike gender-related query terms, race and age-related query terms (e.g. african, asian, elderly, old) yield unclear stereotypical associations. For most race or ethnicity premise words, the words with the highest PMI were uninformative, e.g. \textit{is, the,} and \textit{a}. For age-related queries, the most associated words in entailed hypotheses were synonyms (\textit{senior, older}), and in contradictions were antonyms (\textit{young, children}.)

\begin{figure}[t!]
    \centering
    \includegraphics[width=1\linewidth]{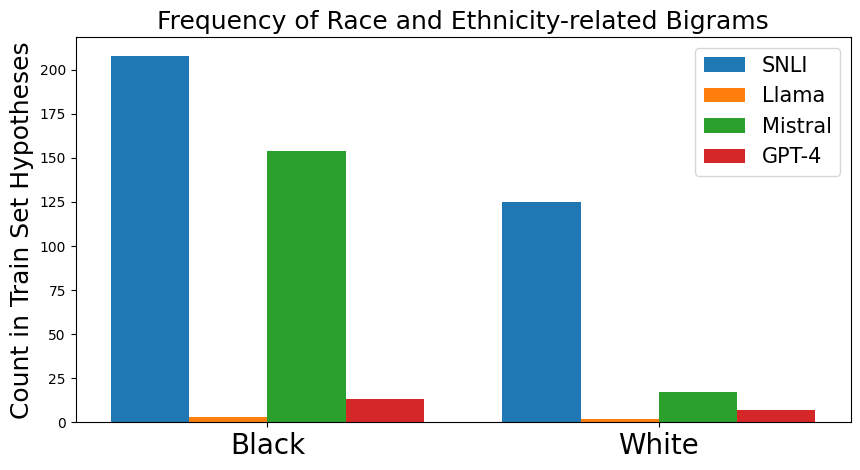}
    \caption{Number of hypotheses in each train set that contain race-related words followed by one of the people-related words from \newcite{rudinger-etal-2017-social}.}
    \label{race-ethnicity}
\end{figure}

Gender-related stereotypical associations seem stronger than racial and ethnic biases in LLM-generated datasets. One possible explanation is that LLM-generated hypotheses typically mention racial and ethnicity-related words much less often than in SNLI's hypotheses, as shown in \autoref{race-ethnicity}.\footnote{In the figure, ``black'' refers to the words \textit{black} and \textit{african}, "white" refers to the words \textit{white} and \textit{european}. The people-related words are the person-related query words from \newcite{rudinger-etal-2017-social}: \textit{woman, man, women, men, girl, boy, girls, boys, female, male, mother, father, sister, brother, daughter, son, person} and \textit{people}.}

\section{Conclusion}
We studied whether Natural Language Inference datasets created by eliciting hypotheses from LLMs contain biases. We used $3$ LLMs to recreate a portion of SNLI and applied standard techniques to determine that like SNLI, LLM-elicited datasets contain annotation artifacts and stereotypical biases. On our LLM-generated NLI datasets, fine-tuned BERT hypothesis-only classifiers achieve between 86-96\% accuracy. Our analyses indicated that LLMs rely on similar strategies and heuristics as crowd-source workers when creating entailed, neutral, and contradicted hypotheses in response to a premise. Our results provide further empirical evidence that well-attested biases in human-elicited text persist in LLM-generated text. Our findings provide a cautionary tale for relying on unfiltered, out-of-the-box LLM-generated textual data for NLP datasets.

\section{Limitations}
 \newcite{srikanth-rudinger-2022-partial} showed that while NLI models \textit{can} gain high performance while ignoring the premise, in practice models still condition on the premise context when making predictions. While our work demonstrated that LLM-elicited datasets can contain biases, 
it is unclear to what extent these biases harm NLI model robustness.

While we aimed to mirror the process used to generate SNLI, our approach is not perfectly comparable. First, SNLI was created by a large pool of crowd-source workers while we focus on just 3 LLMs. Secondly, crowd-source workers could ask clarifying questions, but LLMs could not. Thirdly, the one-shot nature of our prompting prevented LLMs from incorporating instructions across premises, such as the FAQ suggestion to not ``[reuse] the same sentence.''

Another limitation of our work is that we relied on a single prompt to elicit hypotheses from LLMs. Recent work has demonstrated that seemingly insignificant changes to prompts can result in widely varying responses~\cite{mizrahi2023state}. We leave a multi-prompt analysis for future work. 

\section*{Acknowledgments}
We thank anonymous reviewers from current and
past versions of the article for their insightful
comments and suggestions. This research benefited from support by an internal grant awarded by the Bryn Mawr College Faculty Awards and Grants Committee.

\bibliography{anthology,custom}

\appendix

\section{Appendix}

\begin{figure*}[t!]
\begin{framed}
% \small
We will show you the caption for a photo. We will not show you the photo. Using only the caption and what you know about the world:
\begin{itemize}
\item Write one alternate caption that is \textbf{definitely} a \textbf{true} description of the photo. \textit{Example: For the caption ``\textit{Two dogs are running through a field.}'' you could write ``\textit{There are animals outdoors.}"}
\item Write one alternate caption that \textbf{might be} a \textbf{true} description of the photo. \textit{Example: For the caption ``\textit{Two dogs are running through a field.}" you could write ``\textit{Some puppies are running to catch a stick.}"}
\item Write one alternate caption that is \textbf{definitely} a \textbf{false} description of the photo. \textit{Example: For the caption ``\textit{Two dogs are running through a field.}" you could write ``\textit{The pets are sitting on a couch.}" This is different from the} maybe correct \textit{category because it's impossible for the dogs to be both running and sitting.}
\end{itemize}

In response to the original caption, please return the 3 alternate captions in a JSON readable format and include no other commentary.\\\\ \textit{Here is an example of the correct format of response to the prompt:} 

Original caption: "Two dogs are running through a field" 
\\ Three JSON-parseable alternate captions, with "definitely true", "might be true", and "definitely false" descriptions of the photo:\\ \{"true": "There are animals outdoors.",\\"maybe": "Some puppies are running to catch a stick.",\\"false": "The pets are sitting on a couch." \}
\\\\
Now, please generate the 3 alternate captions following the JSON-parseable format described earlier: Original Caption: \textbf{[INSERT SNLI PREMISE]} \\Three JSON-parseable alternate captions, with "definitely true", "might be true", and "definitely false" descriptions of the photo:
\end{framed}

\caption{\label{instructions-1}The prompt provided to all LLMs. The first four paragraphs are identical to those provided to MTurk workers for the SNLI dataset.}
\end{figure*}

\label{sec:appendix}

\begin{comment}
\section{Hyper-params}
\label{app:hyper-params}

\end{comment}

\begin{table*}
\centering
\small
\definecolor{lightgray}{gray}{0.9}
\begin{tabularx}{\textwidth}{|l|X|X|X|}
\hline
Query & \textbf{ENTAIL} & \textbf{NEUTRAL} & \textbf{CONTRA} \\
\hline
\rowcolor{lightgray}
\textbf{african} & \textbf{SNLI}: are is the & \textbf{SNLI}: a to are the is & \textbf{SNLI}: a are is the \\
 & \textbf{GPT-4}: kids a are is person & \textbf{GPT-4}: group of performing a are & \textbf{GPT-4}: a are playing man swimming \\
\rowcolor{lightgray}
 & \textbf{Llama}: a are is people person & \textbf{Llama}: his at are a to & \textbf{Llama}: man group a playing are \\
 & \textbf{Mistral}: an people a in are & \textbf{Mistral}: an or be may are & \textbf{Mistral}: an not and are is \\
\hline
\rowcolor{lightgray}
\textbf{asian} & \textbf{SNLI}: an$^\dagger$ for with near the & \textbf{SNLI}: chinese work up an waiting & \textbf{SNLI}: american$^\ddagger$ white black taking from \\
 & \textbf{GPT-4}: city cooking having food woman & \textbf{GPT-4}: sushi lunch tourists busy exploring & \textbf{GPT-4}: party a$^\ddagger$ men dancing child \\
\rowcolor{lightgray}
 & \textbf{Llama}: students shopping city a$^\ddagger$ food & \textbf{Llama}: cultural individual class restaurant heading & \textbf{Llama}: models$^\dagger$ they astronauts preparing shoot \\
 & \textbf{Mistral}: individual$^\ddagger$ women outdoor an$^\ddagger$ area & \textbf{Mistral}: exploring city tourists$^\ddagger$ an$^\ddagger$ collecting & \textbf{Mistral}: an$^\ddagger$ green outside their cars \\
\hline
\rowcolor{lightgray}
\textbf{asians} & \textbf{SNLI}: are the & \textbf{SNLI}: are the & \textbf{SNLI}: are the \\
 & \textbf{GPT-4}: people are & \textbf{GPT-4}: of & \textbf{GPT-4}: park are \\
\rowcolor{lightgray}
 & \textbf{Llama}: dining$^\ddagger$ people are & \textbf{Llama}: of & \textbf{Llama}: the are \\
 & \textbf{Mistral}: asian$^\ddagger$ are & \textbf{Mistral}: asian$^\ddagger$ are & \textbf{Mistral}: asian$^\ddagger$ are \\
\hline
\rowcolor{lightgray}
\textbf{caucasian} & \textbf{SNLI}: white$^\ddagger$ is & \textbf{SNLI}: is & \textbf{SNLI}: the \\
 & \textbf{GPT-4}: is & \textbf{GPT-4}: is & \textbf{GPT-4}: man is swimming \\
\rowcolor{lightgray}
 & \textbf{Llama}: is & \textbf{Llama}: is & \textbf{Llama}: is \\
 & \textbf{Mistral}: is & \textbf{Mistral}: the is & \textbf{Mistral}: not is the \\
\hline
\rowcolor{lightgray}
\textbf{chinese} & \textbf{SNLI}: is & \textbf{SNLI}: is the & \textbf{SNLI}: a \\
 & \textbf{GPT-4}: are is & \textbf{GPT-4}: a in is & \textbf{GPT-4}: a is in \\
\rowcolor{lightgray}
 & \textbf{Llama}: a is & \textbf{Llama}: someone are is & \textbf{Llama}: a is \\
 & \textbf{Mistral}: is there & \textbf{Mistral}: a be the & \textbf{Mistral}: not is the \\
\hline
\rowcolor{lightgray}
\textbf{indian} & \textbf{SNLI}: the is & \textbf{SNLI}: a to the is & \textbf{SNLI}: on is the \\
 & \textbf{GPT-4}: people is are & \textbf{GPT-4}: is & \textbf{GPT-4}: a is pool swimming \\
\rowcolor{lightgray}
 & \textbf{Llama}: a people is are person & \textbf{Llama}: is & \textbf{Llama}: group a in is \\
 & \textbf{Mistral}: an people is are there & \textbf{Mistral}: a the is & \textbf{Mistral}: the on are is \\
\hline
\end{tabularx}
\caption{Race, Ethnicity, and Nationality-Related Queries}
\end{table*}

\begin{table*}
\centering
\small
\definecolor{lightgray}{gray}{0.9}
\begin{tabularx}{\textwidth}{|l|X|X|X|}
\hline
Query & \textbf{ENTAIL} & \textbf{NEUTRAL} & \textbf{CONTRA} \\
\hline
\rowcolor{lightgray}
\textbf{elderly} & \textbf{SNLI}: old$^\ddagger$ an$^\ddagger$ wearing a are & \textbf{SNLI}: old he a$^\ddagger$ an is & \textbf{SNLI}: old a man at is \\
 & \textbf{GPT-4}: old$^\ddagger$ senior$^\ddagger$ citizen lady instrument & \textbf{GPT-4}: senior$^\ddagger$ old$^\ddagger$ jazz festival musician & \textbf{GPT-4}: young$^\ddagger$ children a playing man \\
\rowcolor{lightgray}
 & \textbf{Llama}: an$^\ddagger$ instrument musical for a & \textbf{Llama}: seniors$^\ddagger$ older$^\ddagger$ citizen$^\ddagger$ senior$^\ddagger$ an & \textbf{Llama}: young$^\ddagger$ child concert woman fashion \\
 & \textbf{Mistral}: seniors$^\ddagger$ older$^\dagger$ an$^\ddagger$ for the & \textbf{Mistral}: older music an$^\ddagger$ a of & \textbf{Mistral}: an$^\ddagger$ a playing is on \\
\hline
\rowcolor{lightgray}
\textbf{old} & \textbf{SNLI}: elderly$^\ddagger$ not a$^\ddagger$ an person & \textbf{SNLI}: hair just home out an & \textbf{SNLI}: young$^\ddagger$ has two a people \\
 & \textbf{GPT-4}: elderly$^\ddagger$ gentleman$^\ddagger$ citizen$^\ddagger$ senior$^\ddagger$ something & \textbf{GPT-4}: citizens$^\dagger$ grandson$^\ddagger$ citizen$^\ddagger$ elderly$^\ddagger$ grandmother$^\ddagger$ & \textbf{GPT-4}: young$^\ddagger$ sandbox her a$^\ddagger$ girl \\
\rowcolor{lightgray}
 & \textbf{Llama}: produce$^\dagger$ elderly$^\ddagger$ woman an$^\dagger$ resting & \textbf{Llama}: elderly$^\ddagger$ citizen$^\ddagger$ senior$^\ddagger$ grandfather an$^\ddagger$ & \textbf{Llama}: young$^\ddagger$ children child$^\ddagger$ her toy \\
 & \textbf{Mistral}: elderly$^\ddagger$ an$^\ddagger$ woman walking a & \textbf{Mistral}: older$^\ddagger$ elderly grandmother$^\dagger$ grandson grandfather & \textbf{Mistral}: elderly$^\ddagger$ young$^\ddagger$ woman an a$^\ddagger$ \\
\hline
\rowcolor{lightgray}
\textbf{teenagers} & \textbf{SNLI}: are the & \textbf{SNLI}: are & \textbf{SNLI}: are the \\
 & \textbf{GPT-4}: young$^\ddagger$ outside people are & \textbf{GPT-4}: high students school game group$^\dagger$ & \textbf{GPT-4}: children$^\ddagger$ library playing are pool \\
\rowcolor{lightgray}
 & \textbf{Llama}: activity engaging people in are & \textbf{Llama}: group of friends are a & \textbf{Llama}: are the \\
 & \textbf{Mistral}: children young people are there & \textbf{Mistral}: could it be are & \textbf{Mistral}: are not the \\
\hline
\rowcolor{lightgray}
\textbf{young} & \textbf{SNLI}: off building jumps a$^\ddagger$ he & \textbf{SNLI}: alone funny high brothers beach & \textbf{SNLI}: kite books birds practicing swims \\
 & \textbf{GPT-4}: children$^\ddagger$ activities physical child$^\ddagger$ a$^\ddagger$ & \textbf{GPT-4}: teenagers test cap giant teenager$^\ddagger$ & \textbf{GPT-4}: snowman adults$^\ddagger$ teenagers old rocking \\
\rowcolor{lightgray}
 & \textbf{Llama}: feature kids$^\ddagger$ sunny observing creative & \textbf{Llama}: teenagers$^\ddagger$ mom skatepark weekend games & \textbf{Llama}: nursing$^\dagger$ seniors$^\ddagger$ citizens elderly senior \\
 & \textbf{Mistral}: shore studying acrobatics children$^\ddagger$ sandy & \textbf{Mistral}: females learning skills siblings school & \textbf{Mistral}: pants kids$^\ddagger$ they toys a$^\ddagger$ \\
\hline
\end{tabularx}
\caption{Age-Related Queries}
\end{table*}

\begin{table*}
\centering
\small
\definecolor{lightgray}{gray}{0.9}
\begin{tabularx}{\textwidth}{|l|X|X|X|}
\hline
Query & \textbf{ENTAIL} & \textbf{NEUTRAL} & \textbf{CONTRA} \\
\hline
\rowcolor{lightgray}
\textbf{boy} & \textbf{SNLI}: boys$^\dagger$ child a$^\ddagger$ his is & \textbf{SNLI}: boys a$^\ddagger$ down trying his & \textbf{SNLI}: girl$^\ddagger$ up asleep a$^\ddagger$ nobody \\
 & \textbf{GPT-4}: active his playground male trick & \textbf{GPT-4}: hide seek kid$^\ddagger$ teenager swimming & \textbf{GPT-4}: kid his$^\ddagger$ girl$^\ddagger$ classroom quietly \\
\rowcolor{lightgray}
 & \textbf{Llama}: child$^\ddagger$ a$^\ddagger$ urban enjoying playing & \textbf{Llama}: young$^\ddagger$ summer person a$^\ddagger$ kid & \textbf{Llama}: surfing teenager suit tie working \\
 & \textbf{Mistral}: a$^\ddagger$ young$^\ddagger$ group child$^\dagger$ standing & \textbf{Mistral}: child$^\ddagger$ how young$^\ddagger$ practicing swimming & \textbf{Mistral}: a$^\ddagger$ man subject$^\ddagger$ reading is \\
\hline
\rowcolor{lightgray}
\textbf{boys} & \textbf{SNLI}: playing are the & \textbf{SNLI}: their and of are a & \textbf{SNLI}: girls$^\ddagger$ playing are$^\dagger$ the \\
 & \textbf{GPT-4}: children$^\ddagger$ sport event activity participating & \textbf{GPT-4}: kids$^\ddagger$ game their playing group & \textbf{GPT-4}: are$^\ddagger$ beach a swimming the \\
\rowcolor{lightgray}
 & \textbf{Llama}: children$^\ddagger$ physical activity$^\dagger$ engaging$^\dagger$ outdoors & \textbf{Llama}: sport kids$^\ddagger$ team participating game & \textbf{Llama}: players competing team game astronauts \\
 & \textbf{Mistral}: children$^\ddagger$ event sport outdoors playing & \textbf{Mistral}: children$^\ddagger$ sport running kids$^\ddagger$ fun & \textbf{Mistral}: kids$^\ddagger$ photo inside individuals in \\
\hline
\rowcolor{lightgray}
\textbf{girl} & \textbf{SNLI}: girls her a$^\ddagger$ child wearing & \textbf{SNLI}: girls she a$^\ddagger$ plays her$^\dagger$ & \textbf{SNLI}: she guy boy a$^\ddagger$ wearing \\
 & \textbf{GPT-4}: female$^\ddagger$ a$^\ddagger$ riding musical instrument & \textbf{GPT-4}: woman$^\ddagger$ young$^\ddagger$ teenager lady$^\ddagger$ child & \textbf{GPT-4}: boy$^\ddagger$ video a$^\ddagger$ his climbing \\
\rowcolor{lightgray}
 & \textbf{Llama}: a$^\ddagger$ wearing place public the & \textbf{Llama}: instrument expressing woman$^\ddagger$ young$^\ddagger$ favorite & \textbf{Llama}: ice child$^\ddagger$ professional mountain toy \\
 & \textbf{Mistral}: wearing a$^\ddagger$ young physical activity & \textbf{Mistral}: woman$^\ddagger$ young$^\ddagger$ subject little her & \textbf{Mistral}: a$^\ddagger$ any wearing subject book \\
\hline
\rowcolor{lightgray}
\textbf{girls} & \textbf{SNLI}: girl some$^\ddagger$ their wearing are & \textbf{SNLI}: girl some they at are & \textbf{SNLI}: boys$^\ddagger$ their two playing a \\
 & \textbf{GPT-4}: females$^\ddagger$ children$^\ddagger$ game sport participating & \textbf{GPT-4}: match group team a$^\ddagger$ practicing & \textbf{GPT-4}: boys$^\ddagger$ field studying football soccer \\
\rowcolor{lightgray}
 & \textbf{Llama}: students athletes indoors activity physical & \textbf{Llama}: teenagers$^\dagger$ teammates women$^\dagger$ friendly sisters & \textbf{Llama}: celebrities$^\ddagger$ premiere cats movie show \\
 & \textbf{Mistral}: sports celebrating people$^\ddagger$ are$^\ddagger$ there$^\dagger$ & \textbf{Mistral}: females$^\ddagger$ female children$^\dagger$ athletes could$^\dagger$ & \textbf{Mistral}: children individuals$^\ddagger$ a park are \\
\hline
\rowcolor{lightgray}
\textbf{female} & \textbf{SNLI}: woman$^\ddagger$ a is the & \textbf{SNLI}: woman$^\ddagger$ wearing a in is & \textbf{SNLI}: male$^\ddagger$ woman playing a is \\
 & \textbf{GPT-4}: woman$^\ddagger$ athlete the playing performing & \textbf{GPT-4}: woman$^\dagger$ practicing lady her a & \textbf{GPT-4}: skiing basketball mountain man a \\
\rowcolor{lightgray}
 & \textbf{Llama}: a playing person is & \textbf{Llama}: woman$^\ddagger$ a of is & \textbf{Llama}: fashion man playing a is \\
 & \textbf{Mistral}: woman$^\ddagger$ performing an playing is & \textbf{Mistral}: exercise woman a be for & \textbf{Mistral}: a subject playing person is \\
\hline
\rowcolor{lightgray}
\textbf{he} & \textbf{SNLI}: man a$^\dagger$ & \textbf{SNLI}: a & \textbf{SNLI}: man a \\
 & \textbf{GPT-4}: man$^\ddagger$ wearing a$^\dagger$ in person & \textbf{GPT-4}: in his a & \textbf{GPT-4}: a pool in swimming \\
\rowcolor{lightgray}
 & \textbf{Llama}: wearing a in person & \textbf{Llama}: in for a the & \textbf{Llama}: cooking pool swimming at a \\
 & \textbf{Mistral}: a in person & \textbf{Mistral}: someone a for be & \textbf{Mistral}: a person not \\
\hline
\rowcolor{lightgray}
\textbf{male} & \textbf{SNLI}: man a people outside is & \textbf{SNLI}: practicing man$^\ddagger$ from his a & \textbf{SNLI}: waiting an man his sitting \\
 & \textbf{GPT-4}: man$^\ddagger$ at performing a two & \textbf{GPT-4}: man$^\dagger$ a$^\dagger$ park at on & \textbf{GPT-4}: skiing a mountain woman cooking \\
\rowcolor{lightgray}
 & \textbf{Llama}: their a outdoors on is & \textbf{Llama}: man$^\ddagger$ practicing break couple on & \textbf{Llama}: sunny preparing man an park \\
 & \textbf{Mistral}: man performing space riding a & \textbf{Mistral}: man$^\ddagger$ a$^\dagger$ performing his couple & \textbf{Mistral}: subject a wearing bench sitting \\
\hline
\end{tabularx}
\caption{Additional Gender-Related Queries}
\end{table*}

\end{document}